\pdfoutput=1

\documentclass[11pt]{article}

\usepackage[preprint]{acl}

\usepackage{times}
\usepackage{latexsym}
\usepackage[inkscapelatex=false]{svg}
\usepackage[T1]{fontenc}

\usepackage[utf8]{inputenc}

\usepackage{microtype}

\usepackage{inconsolata}

\usepackage{graphicx}
\usepackage{mwe} 
\usepackage[utf8]{inputenc} 
\usepackage[T1]{fontenc}    
\usepackage{hyperref}       
\usepackage{url}            
\usepackage{booktabs}       
\usepackage{amsfonts}       
\usepackage{nicefrac}       
\usepackage{microtype}      
\usepackage{xcolor}         
\usepackage{amsmath} 
\usepackage{multirow}
\usepackage{subcaption}
\usepackage{caption}
\usepackage[colorinlistoftodos,bordercolor=orange,backgroundcolor=orange!20,linecolor=orange,textsize=scriptsize]{todonotes}

\title{Continual Quantization-Aware Pre-Training: When to transition from 16-bit to 1.58-bit pre-training
for BitNet language models?}

\author{Jacob Nielsen \and Peter Schneider-Kamp \and Lukas Galke \\
Department of Mathematics and Computer Science, University of Southern Denmark\\ Odense, Denmark \\ \{jacn, galke, petersk\}@imada.sdu.dk}

\begin{document}
\maketitle
\begin{abstract}
Large language models (LLMs) require immense resources for training and inference. Quantization, a technique that reduces the precision of
model parameters, offers a promising solution for improving LLM efficiency and sustainability. While post-training quantization methods
typically achieve 4-8 bits per parameter, recent research suggests that training LLMs with 1.58 bits per weight parameter from scratch can maintain model accuracy while greatly reducing memory requirements and energy consumption at inference time.
Here, we investigate a training strategy for
quantization-aware pre-training, where the models are first trained with 16-bit precision and then transition into 1.58-bit quantization-aware training. Our results on 11 downstream tasks show that this 16-to-1.58-bit training strategy is preferable over full 1.58-bit training and leaves models closer to those which have undergone 16-bit training. We further investigate the effects of retaining the optimizer state at the transition point and gradually phasing in quantization strength -- finding that both techniques alleviate the magnitude of loss spikes, but also that these effects can be compensated through further training.
\end{abstract}

\section{Introduction}

Large Language Models (LLMs) have revolutionalized natural language processing  and are making a strong entry into both related and unrelated industries. However, deployment of LLMs comes with a series of obstacles such as memory-usage, latency, and throughput. Environmental considerations regarding energy consumption of both training and inference becomes an increasingly important aspect~\cite{schwartz2020green,DBLP:conf/aaai/StrubellGM20}. 

Quantization-aware training of language models, i.e. preparing the model for later quantization already during training, has shown promising results~\cite{wang2023bitnet,ma2024era}. However, the resulting models tend to require more parameters to compensate for the reduction in bit-precision per parameter~\cite{kumar2025scaling, nielsen2024bitnetb158reloadedstateoftheart}.
Here, we investigate a strategy of first training with standard precision in an initial phase, followed by a second phase of 1.58-bit quantization-aware training, where weights are being quantized to either -1, 0, or 1.

Reducing inference compute demands of language models is particularly critical in the context of recent advances in scaling test time compute~\cite{guo2025deepseek,muennighoff2025s1simpletesttimescaling,jaech2024openai}. If the trend of scaling test-time compute continues, we can assume that inference will soon dominate the resource consumption in a language model's life cycle. 

Recent works in 1-bit~\citep{wang2023bitnet} and 1.58-bit quantization-aware training \cite{ma2024era, nielsen2024bitnetb158reloadedstateoftheart, bitsenoughbottomup}, demonstrate the potential of training in 1.58-bit precision while retaining most of the performance, mitigating some of the drawbacks of existing post-training quantization techniques in both NLP and computer vision~\citep{frantar2023gptq,li2023vit}.
To achieve competitive model performance, these quantization-aware training strategies keep 16-bit-precision weights at training time (``shadow weights''), which are quantized on-the-fly to 1-bit or 1.58-bit precision during forward passes. Straight-through estimated gradients~\cite{straightthrough} are then used to update the shadow weights.

While such a training strategy initially requires more memory and compute than pre-training a regular language model, its benefits can be harvested after training. At inference time, the final shadow weights can be quantized once and for all, after which the shadow weights can be discarded, yielding a model with a 8-16 times reduced memory footprint. With appropriate kernels, this effectively allows us to replace costly matrix multiplications within linear layers with computationally more efficient integer addition operations.
 
Although these methods provides an exciting avenue for efficient inference on models trained with 1.58-bit quantization-aware training techniques, existing models are not eligible for efficient inference out of the box~\citep{wang2023bitnet}. This is increasingly important as increasing model size implies corresponding resource requirements, making training from scratch both resource heavy and environmentally challenging. 

Recent analysis have hinted at 7-8 bits being compute optimal~\cite{kumar2025scaling}. Pushing the bit-representation even lower currently requires a quantization-aware training strategy as proposed in BitNet~\cite{wang2023bitnet,ma2024era}.

Concurrent work on quantization-aware training has shown that it is generally possible to convert a 16-bit model and into a 1.58-bit model through continual pre-training~\cite{huggingfaceFinetuningLLMs}. However, the experiments reported upon only consider the case of starting with an already pre-trained language model and do not compare against training with low precision from scratch. In related matter, recent work has shown that ``overtrained'' networks, such as fully-pre-trained language models, do not represent an ideal starting point for quantization \cite{kumar2025scaling}.

\begin{figure}[t]
    \centering
    \includegraphics[width=\linewidth]{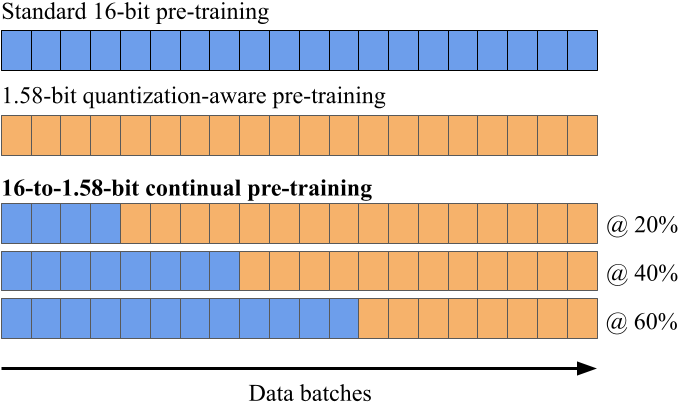}
    \caption{16-to-1.58-bit continual pre-training. Blue and yellow denotes batches processed under 16-bit and 1.58-bit training, respectively.}
    \label{fig:one}
\end{figure}

It is so far unclear which strategy to choose when pre-training from scratch: Is it preferable to train with 1.58-bit quantization-aware training all the way, or is it preferable to start with standard 16-bit training and then transition into 1.58-bit quantization-aware training (see Figure~\ref{fig:one}). And if so, when should we switch from 16-bit to 1.58-bit training? Is a it beneficial to gradually phase in the quantization strength? And what role does the retention of the optimizer state (if available) play?

In this work, we investigate these questions to shed new light on the potential of continuing a pre-training run in 1.58-bit. We aim to provide insights into the potential of converting existing pre-trained models to 1.58-bits and into developing more efficient ways of pre-training new models. Ultimately, this work contributes to a future with more efficient and environmentally-friendly LLMs.

In order to answer these questions and achieve our aims, we systematically perform a number of experiments to identify a data-optimal transition point from 16-bit to 1.58-bit training. We further investigate whether it makes sense to phase in the quantization strength and whether the optimizer state is of importance, providing a basis for future work on expand upon and for practitioners to receive guidance in converting existing models and pre-training new models.

Challenging both conventional wisdom and common-sense expectations, our systematic comparison reveals that models that are first trained with 16-bit precision, and only later transition into 1.58-bit quantization-aware training, are more ultimately effective than models that are fully trained with 1.58-bit quantization-aware training.

In addition, we find that, while retaining the optimizer state indeed helps to alleviate a temporary increase in training loss, continual pre-training with a fresh optimizer state attains similar loss values after a limited number of optimization steps.

Furthermore, we find that gradually introducing quantization strength provides little to no benefits. A continual pre-training with full quantization strength temporarily increases training loss, but attains similar loss values, and also downstream performance, after a limited number of steps.

These results together indicate that the availability of optimizer state is not an obstacle when continuing pre-training from openly available language models and that no new hyperparameters (regarding quantization strength) are needed.
In summary, our contributions are as follows:
\begin{itemize}
    \item Systematic comparison between 1.58-bit pre-training from scratch and hybrid 16-bit training with 1.58-bit continual pre-training.
    \item Results from 11 downstream tasks showing that  16-to-1.58-bit continual pre-training is more effective than full 1.58-bit training.
    \item An analysis revealing limited effects of the gradual phasing in of quantization strength and the retention of optimizer states.
\end{itemize}

\section{Related work}

\paragraph{Early attempts of ternary-weight neural nets.}
The exploration into ternary weights was motivated by finding a better trade-off
between accuracy and complexity than binary neural networks, which had suffered
a substantial decrease in performance, effectively hindering the usability of such networks
\citep{chen2021fatnn}. 
Earlier attempts of binary- and ternary-weight neural networks showcased ternary superiority over binary weights, while showing promising results in the computer vision domain employing a direct optimization of the quantization \citep{li2016ternary, zhu2016trained}.

\paragraph{Post-training quantization.}
The most common approaches for quantization fall under the category of post-training quantization. Those include approaches such as Generative Pre-trained Transformer Quantization (GPTQ) \citep{frantar2023gptq} and Activation-aware Weight Quantization (AWQ) \citep{lin2024awq}. A similar proposal for the vision transformer ~\citep{li2023vit} represents one of many efforts in the computer vision domain. Post-training quantization approaches come with an inherent decrease in performance, trading increased latency and throughput and decrease memory for precision~\cite{kumar2025scaling}.

\paragraph{Quantization-aware training.}
Quantization-aware training was already proposed in earlier work on post-training such as LLM-QAT ~\citep{liu2023llmqat} and QA-LoRA ~\citep{xu2023qaloraquantizationawarelowrankadaptation}. These methods directly optimize the quantized weights with respect to an objective function such that there is no decrease in performance when the model is used for inference.  

Recently, we have seen a number of works on 1-bit~\citep{wang2023bitnet} and 1.58-bit~\citep{ma2024era} quantization-aware techniques demonstrating strong performance in LLM performance, yielding a small or no loss in precision depending on model sizes. Other works have demonstrated strong potential for multi-modal architectures \citep{sundaram2024llavaolmobitnet1b} and spiking language models ~\citep{bal2024exploring}. 
Lastly, we have seen an investigation into the potential of 1.58-bit in small language and vision models and the definition of a scaling law for decoder-only models for reintroducing capacity~\citep{nielsen2024bitnetb158reloadedstateoftheart}. Similar scaling laws hold for encoder-only models while encoder-decoder models seem to be less predictable \citep{bitsenoughbottomup}. This latest work further shows that also non-transformer models, such as plain multi-layer perceptions and graph neural networks can attain similar performance as their 16/32-bit counterparts, even without increasing the number of parameters.

\paragraph{Summary.}
Research on language model quantization shows promising results but is also limited by the effectiveness of the final models compared to standard precision models. There is a strict distinction between post-training methods and quantization-aware training. So far, it remains unexplored whether an initial phase of standard 16-bit precision training would improve or worsen the performance of the final model when continuing with 1.58-bit quantization-aware pre-training.

\section{Methods}\label{sec:methods}

We first recapitulate the basics of quantization-aware training (Section~\ref{sub:methods:qat}) before describing the proposed strategy of continual 1.58-bit pre-training (Section~\ref{sub:methods:cpt-1.58}) and discussing critical considerations concerning optimizer states (Section~\ref{sub:methods:optstate}) and gradually phasing in quantization strength (Section~\ref{sub:methods:warmup}).

\subsection{Background on 1.58-bit training}\label{sub:methods:qat}
Recent work has focused on specifically on 1.58-bit quantization-aware training techniques \cite{wang2023bitnet, nielsen2024bitnetb158reloadedstateoftheart, bitsenoughbottomup}. Specifically, these works investigate ternary networks, where the weights only can take on the values -1, 0 and 1. The quantization is guided by 16-bit precision ``shadow weights'', which are quantized during the forward passes and rely on a straight-through estimator for optimization.  Activations are also commonly quantized in a given range $Q_b$, which is then multiplied with the quantized weight-matrix. Intuitively, the the weight-matrix then either is adding, ignoring, or subtracting the previous layer's activations using -1, 0, or 1, respectively. 

The key idea of 1.58-bit training is to maintain 16-bit precision ``shadow weights'' and quantize them on-the-fly. We follow the basic formulation of BitNet~\cite{wang2023bitnet} for replacing activations and weights in all linear layers of a language model as follows: 

{\small \begin{align*}
   \mathbf{W}_\mathrm{quant} = &\max(-1, \min(1, \operatorname{round}(\mathbf{W} \cdot w_\mathrm{scale}))\\
\mathbf{x}_\mathrm{quant} =  &\max(-Q_b,\min(Q_b-1, \operatorname{round}(\hat{\mathbf{I}} \cdot x_\mathrm{scale}))
\end{align*}}
where $\mathbf{W}$ denotes the weight parameters of the linear layers (``shadow weights'') and $\hat{\mathbf{I}}$ denotes the normalized input. $Q_b$ defines the integer range of the activations, which defaults to $8$ bits, i.e., $256$ possible values. 
The scaling factors $w_\mathrm{scale}$ and $x_\mathrm{scale}$ are derived from the respective means of the weight matrix $\mathbf{W}$ and the layer's input $\hat{\mathbf{I}}$. 

\subsection{Continual 1.58-bit Pre-training}\label{sub:methods:cpt-1.58}
We hypothesize that, to obtain the best possible 1.58-bit model given a fixed amount of data, the model needs to first find a good set of 16-bit parameters before those can be quantized to 1.58-bit. As such, we hypothesize that there exists a point $t^\star$ during training, at which one can  switch from 16-bit training to 1.58-bit training in order to achieve an ultimately better 1.58-bit model, than one would obtain by complete quantization-aware training on the same data.

Formally, given a training set $\left\{ \mathbf{x_i} \right\}_{i<N}$, we hypothesize that there exists a specific point in training $t^\star$, such that a model first trained with 16 bit on $\mathbf{x_0}, \mathbf{x_1}, \ldots, \mathbf{x_{t^\star}}$ and then trained with 1.58 bit on $\mathbf{x_{{t^\star}+1}}, \mathbf{x_{{t^\star}+2}} \ldots, \mathbf{x_{N-1}}$ ultimately performs better than a model pre-trained with 1.58-bits on the full training set $\mathbf{x_0}, \mathbf{x_1} \ldots \mathbf{x_{N-1}}$. See Figure~\ref{fig:one}.
At the transition point $t^\star$, the 16-bit weights turn into ``shadow weights'' for quantization-aware training.

\subsection{Optimizer State Retention}\label{sub:methods:optstate}
A key consideration in investigating continual 1.58-bit pre-training is the availability of optimizer states. If the optimizer states are available, we expect a smooth transition from pre-training to continued pre-training.
In our strictly controlled experimental setup, the optimizer states are available.
However, in practice, optimizer states cannot be assumed to be available, when one would seek continuing pre-training on arbitrary base models. We take the opportunity to investigate the effect of optimizer states being available vs.\ assuming that they were not available.

\subsection{Phasing in Quantization Strength}\label{sub:methods:warmup}
We further investigate a recently proposed technique to gradually increase quantization strength. In this formulation, we would train a model on $\mathbf{x_0}, \mathbf{x_1}, \ldots, \mathbf{x_s}$ with 16-bit training, and then gradually introduce quantization strength on the data $\mathbf{x_{s+1}}, \mathbf{x_{s+2}}, \ldots \mathbf{x_{t^\star}}$ \citep{huggingfaceFinetuningLLMs}.
Specifically, we consider an extra hyperparameter $\lambda$ integrated into the calculation of $\mathbf{x}_\mathrm{quant}$ and $\mathbf{w}_\mathrm{quant}$ denoting the activations and weight quantization respectively. We formulate the gradual quantization strength as:

{\small \begin{align*}
\mathbf{x}_\text{softquant} &= \mathbf{x} + \lambda \cdot \operatorname{detach}\left( \mathbf{x}_\mathrm{quant} - \mathbf{x} \right)\\
\mathbf{W}_\text{softquant} &= \mathbf{W} + \lambda \cdot \operatorname{detach} \left( \mathbf{W}_\mathrm{quant} - \mathbf{W}\right)\\
\end{align*}
}
where $\lambda$ is a hyperparameter that controls the quantization strength, and $\operatorname{detach}(\cdot)$ prevents gradient flow by detaching the operations from the computation graph. 
Reflecting on previous work \citep{huggingfaceFinetuningLLMs} and experimenting with different schedules, we observed that the main effect of using a gradual phasing in of quantization strength seems to happen in the transition from near-full quantization to full quantization, i.e., for high values of $\lambda$. In our experiments, the value for the hyperparameter $\lambda$ at each optimization step is therefore determined by the following schedule:

{\small $$
\lambda(t) =
\begin{cases}
    0, & \text{if }{t \leq s}\\
2\sigma(\frac{5(t-s)}{t^\star-s})-1,  & \text{if } s < t \leq {t^\star}\\
    1, & \text{otherwise}\\
\end{cases} 
$$}

Here, $\sigma$ is the logistic sigmoid function:
{\small $$
\sigma(x) = \frac{1}{1 + e^{-x}}
$$}
This schedule is based on the shifted and scaled part of the sigmoid function for non-negative $x$.

\section{Experimental Setup}
We conducted experiments investigating the potential of continuing the pretraining in 1.58-bits from (partially) pre-trained 16-bit models, initializing the 1.58-bit model's ``shadow-weights'' with the pre-trained 16-bit weights. Furthermore, we investigate the impact of optimizer state retention and phasing in quantization strength.

\paragraph{Architecture}
Specifically, we employ the model architecture of Open Language Models \citep{OLMo} in their official 1B parameters configuration.
We use the BitLinear package\footnote{https://pypi.org/project/bitlinear/} to all replace all \texttt{nn.Linear} layers within an OLMo model by \texttt{BitLinear} layers (as described in Section~\ref{sec:methods}), which includes both projection layers within the attention and feed-forward modules. 

\paragraph{Training}
We train all models on the Dolma dataset~\cite{dolma}, with OLMo's standard hyperparameters\footnote{https://github.com/allenai/OLMo/blob/main/configs/official-0724/OLMo-1B.yaml}. 
In particular, optimization is carried out by the AdamW optimizer \citep{kingma2014adam} with a cosine learning rate scheduler with warmup. We employ a sequence length of 2048 and a batch size of 2048, totaling to a batch size of 4M tokens. 
All experiments are run for a total of $10{,}000$ optimizer steps.

\paragraph{Experimental Conditions}
In this setting, we compare various conditions:
\begin{itemize}
    \item We vary the point in training, where we transition from 16 bit to 1.58 bit precision: either at 2K, 4K, or 6K steps.
    \item We control the retention of optimizer states: keeping the optimizer state alive vs. resetting the optimizer.
    \item We compare a sharp transition from 16-bit training to 1.58-bit training against a soft transition by gradually phasing-in quantization strength.
\end{itemize}

\paragraph{Baselines}
As our baselines, we consider training the model entirely under 1.58-bit quantization-aware training (for brevity: full 1.58-bit training) and entirely under standard 16-bit training (full 16-bit training). The expectation is that the performance of continually pre-trained 1.58-bit models will land between those two baselines, with full 16-bit training being expected to perform best.

\paragraph{Evaluation}
For downstream evaluation, we employ ARC-easy~\citep[reasoning][]{clark2018think}, CommitmentBank~\cite{de2019commitmentbank}, COPA~\cite{roemmele2011choice}, HellaSwag~\cite{zellers2019hellaswag}, MRPC~\cite{dolan2005automatically}, OpenBookQA~\cite{OpenBookQA2018}, PIQA~\cite{Bisk2020}, RTE~\cite{dagan2005pascal}, SciQ~\cite{SciQ}, SST-2~\cite{socher-etal-2013-recursive}, and WinoGrande~\cite{ai2:winogrande}.
All downstream task evaluations are carried out via zero-shot inference, i.e., without fine-tuning and relying solely on pre-trained knowledge and generalization capabilities.

\section{Results}
We first present the results regarding 16-to-1.58-bit quantization-aware training in Section~\ref{sub:results:cpt}, followed by results comparing effects of and optimizer state retention (Section~\ref{sub:results:optstate}) and of phasing in of quantization strength (Section~\ref{sub:results:quantwarmup}).
Lastly, we report the results of the evaluation on downstream tasks in Section~\ref{sub:results:downstream}.

\begin{figure*}[ht]
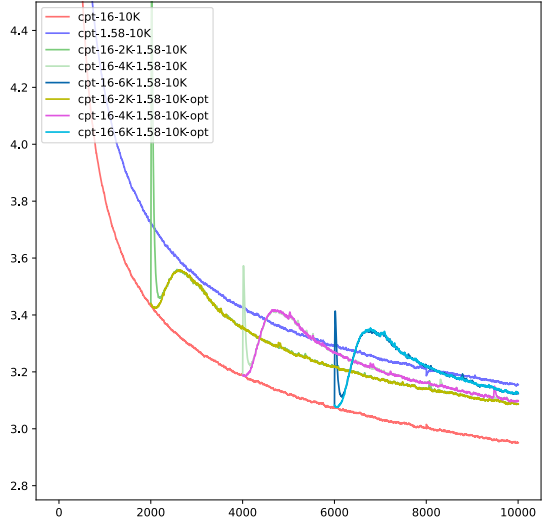

    \centering
    \caption{Training loss curves comparing the effect of different variants of 1.58-bit continual pre-training from 16-bit models into 1.58-bit models. Our baseline is full 1.58-bit quantization-aware pre-training (cpt-1.58-10K; blue). Standard 16-bit training without quantization (cpt-16-10K) is displayed in red. Continual pre-trainings, i.e., transitioning from 16-bit into 1.58-bit training, are marked according to the transition points at 2K, 4K, and 6K optimizer steps, respectively. All models have been trained for 10K steps in total.
    All training loss curves have been smoothed with an exponential filter with window size $64$.}
    \begin{subfigure}[b]{0.45\textwidth}
        \centering
        \includesvg[width=\textwidth]{plots/cpt-reloaded-opt-train.svg}
        \caption{Comparing the 16-bit training with different stages of 1.58-bit continued pretraining and investigating the impact of transferring the optimizer-states.
        Training runs marked with the ``\texttt{-opt}'' suffix have their optimizer state retained.
        }
        
        \label{fig:cpt-opt}
    \end{subfigure}
    \hfill
    \begin{subfigure}[b]{0.45\textwidth}
        \centering
        \includesvg[width=\textwidth]{plots/cpt-reloaded-train.svg}
        \caption{Comparing the 16-bit training with different stages of 1.58-bit continued pretraining and investigating the impact of phasing in quantization.
        Training runs marked with the ``\texttt{-nowarm}'' suffix do not employ phasing in.}
        \label{fig:cpt-warm-act}
    \end{subfigure}
    \label{fig:cpt-train-loss}
\end{figure*}

\subsection{Results for Continual 1.58-bit Pre-training}\label{sub:results:cpt}

We conduct experiments investigating the potential in continuing the pre-training in 1.58-bits from a 16-bit models, initializing the 1.58-bit model's ``shadow weights'' with the pre-trained 16-bit weights. Further, we investigate the impact of gradually phasing in the quantization strength, and, lastly, continuing the 1.58-bit pre-training  with the 16-bit pre-training optimizer.

We show that the best regime encountered in these pre-trainings consists of 2K 16-bit steps, demonstrating a gradually decrease in performance for both 4K and 6K 16-bit steps starting-points. Even more importantly, all three continual 1.58-bit training runs achieve a better performance than the full 1.58-bit training, demonstrating that training 1.58-bit models from scratch is suboptimal.

In Figure \ref{fig:cpt-opt}, we observe that training a 16-bit model from scratch for 10K steps yields a loss value at $2.95$ (red), whereas a fully 1.58-bit training achieves a loss around $3.15$ (purple) after equally many steps. Continuing the pre-training from 2K 16-bit steps yields the best of the continued pre-training variants, with a resulting loss value around $3.088$ (green). We observe that transitioning early yields a spike in the loss for around $100$ steps before the model is able to catch up. 
Continuing from 4K 16-bit steps (light green) exhibits a smaller curve-spike, yielding a loss value at $3.097$ after 10K steps. 
Last, continuing from 6K 16-bit steps (dark blue) exhibits an even smaller curve-spike yielding a loss value of around $3.12$.

Overall, it is clear that training 1.58-bit from scratch is suboptimal. Furthermore, the training loss curves indicate that 16-bit training for more steps is likely detrimental to the overall training loss. The choice of 2K 16-bit steps seems to be optimal within the four regimes considered.

\subsection{Effect of Optimizer State Retention}\label{sub:results:optstate}
We investigate the impact of retaining the optimizer from the 16-bit pre-training, a warm optimizer, instead of constructing a newly initialized optimizer. We observe that spikes in loss curves in Figure \ref{fig:cpt-opt} can be dampened by employing the optimizer states from the 16-bit training, demonstrated by the experiments with yellow, pink, and teal loss curves, yielding a lower and smoother spike-curve across all transfers. However, we observe the smoothening in the transition is gradually having a smaller effect, when transitioning at later stages (4K and 6K steps) of the training. More importantly, the warm-optimizer runs resulted in comparable loss values to their corresponding cold-optimizer runs.

\subsection{Effect of Phasing in Quantization Strength}\label{sub:results:quantwarmup}
Figure \ref{fig:cpt-warm-act} shows the impact of phasing in quantization strength. In the ``\texttt{nowarm}'' case, the training loss exhibits large spikes, which are however recovered relatively soon. After 10K total steps, the training loss is again at the same level  compared to  gradually phasing in quantization strength. This suggests that phasing in quantization strength is not having a lasting impact when pre-training language models, and that there is sufficient time to recover from any disturbance imposed by the abrupt quantization. 
It seems that quantization may as well be introduced at one point in time, alleviating the need for tuning extra hyperparameters and schedules determining the quantization strength.

\begin{table*}[ht]
    \centering
    \tiny
    \begin{tabular}{lccccccccccc}
    \toprule
        \textbf{Model} & \textbf{ARC-easy} & \textbf{CommitB} & \textbf{COPA} & \textbf{HellaSwag} & \textbf{MRPC} & \textbf{OpenBookQA} & \textbf{PIQA} & \textbf{RTE} & \textbf{SciQ} & \textbf{SST-2} & \textbf{WinoGrande} \\
         \midrule 
         full 1.58-bit training & 0.4561 & 0.4107 & 0.6300 & 0.3212 & \textit{0.8122} & \textit{0.2760} & 0.6425 & 0.5343 & \textbf{0.6840} & \textit{0.5803} & \textit{0.5185} \\
         full 16-bit training & \textit{0.4596} & 0.4107 & \textit{0.6500} & \textbf{0.3607} & \textit{0.8122} & \textit{0.2760} & \textit{0.6589} & 0.5379 & 0.6780 & 0.5447 & 0.5170 \\
         \midrule
         cpt-16-2K-1.58-10K & 0.4526 & \textit{0.4464} & \textbf{0.6900} & 0.3375 & 0.8105 & 0.2700 & \textbf{0.6621} & 0.4874 & 0.6760 & 0.5539 & 0.5146 \\
         cpt-16-2K-1.58-10K-nowarm & 0.4456 & 0.4107 & 0.6300 & 0.3342 & \textbf{0.8134} & 0.2700 & 0.6480 & \textit{0.5487} & 0.6760 & \textbf{0.6101} & 0.4988 \\
         cpt-16-2K-1.58-10K-opt & \textbf{0.4632} & \textbf{0.5179} & 0.6300 & \textit{0.3376} & 0.7951 & 0.2620 & 0.6420 & 0.4946 & \textit{0.6830} & 0.4989 & \textbf{0.5264} \\
         
         cpt-16-4K-1.58-10K & 0.4544 & 0.4286 & \textit{0.6500} & 0.3328 & \textit{0.8122} & 0.2680 & 0.6458 & 0.5235 & 0.6520 & 0.5229 & 0.5146 \\
         cpt-16-4K-1.58-10K-nowarm & 0.4491 & 0.4286 & 0.6300 & 0.3329 & \textit{0.8122} & 0.2560 & 0.6415 & \textbf{0.5668} & 0.6660 & 0.5745 & 0.5107 \\
         cpt-16-4K-1.58-10K-opt & 0.4439 & 0.4107 & 0.6200 & 0.3339 & \textit{0.8122} & \textbf{0.2780} & 0.6491 & 0.5451 & 0.6600 & 0.5092 & 0.4996 \\
         cpt-16-6K-1.58-10K & 0.4351 & 0.3929 & 0.6000 & 0.3262 & \textit{0.8122} & 0.2560 & 0.6366 & 0.5451 & 0.6790 & 0.5791 & 0.4838 \\
         cpt-16-6K-1.58-10K-nowarm & 0.4491 & 0.4286 & 0.6100 & 0.3240 & \textit{0.8122} & 0.2640 & 0.6398 & \textit{0.5487} & 0.6700 & 0.5183 & 0.4988 \\
         cpt-16-6K-1.58-10K-opt & 0.4333 & 0.3750 & 0.6100 & 0.3291 & \textit{0.8122} & \textit{0.2760} & 0.6360 & 0.5090 & 0.6810 & 0.5493 & 0.5107 \\
         \bottomrule
    \end{tabular}
    \caption{Final downstream evaluation with the best results marked in \textbf{bold} and the runner-ups in \textit{italics}. All model variants have been trained for 10K steps in total.}\label{tab:final}
\end{table*}

\subsection{Results of the Downstream Evaluation}\label{sub:results:downstream}
We evaluated the downstream performance on all downstream tasks after each 1K optimizer steps. 

\begin{figure*}[hp!]
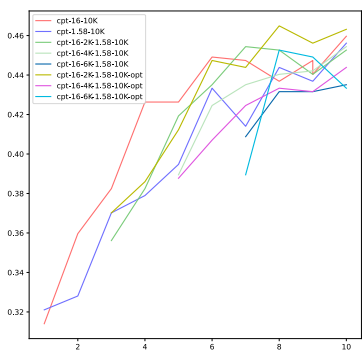

    \centering
    \caption{Downstream evaluation of full 16-bit, continually pre-trained 1.58-bit, and full 1.58 trainings. The y axes are not scaled such that relative differences are more visible.}
    \label{appendix:downstream}
    \begin{subfigure}[b]{0.3\textwidth}
        \centering
        \includesvg[width=\textwidth]{downstreamplots/cpt-reloaded-opt-eval__downstream__arc_easy_acc.svg}
        \caption{ARC-Easy~\citep{clark2018think}}
        \label{appendix:downstream:a}
    \end{subfigure}
    \hfill
    \begin{subfigure}[b]{0.3\textwidth}
        \centering
        \includesvg[width=\textwidth]{downstreamplots/cpt-reloaded-opt-eval__downstream__commitment_bank_acc.svg}
        \caption{CommitmentBank~\citep{de2019commitmentbank}}
        \label{appendix:downstream:b}
    \end{subfigure}
    \hfill
    \begin{subfigure}[b]{0.3\textwidth}
        \centering
        \includesvg[width=\textwidth]{downstreamplots/cpt-reloaded-opt-eval__downstream__copa_acc.svg}
        \caption{COPA~\citep{roemmele2011choice}}
        \label{appendix:downstream:c}
    \end{subfigure}

    \begin{subfigure}[b]{0.3\textwidth}
        \centering
        \includesvg[width=\textwidth]{downstreamplots/cpt-reloaded-opt-eval__downstream__hellaswag_len_norm.svg}
        \caption{HellaSwag \citep{zellers2019hellaswag}}
        \label{appendix:downstream:d}
    \end{subfigure}
    \hfill
    \begin{subfigure}[b]{0.3\textwidth}
        \centering
        \includesvg[width=\textwidth]{downstreamplots/cpt-reloaded-opt-eval__downstream__mrpc_f1.svg}
        \caption{MRPC \citep{dolan2005automatically}}
        \label{appendix:downstream:e}
    \end{subfigure}
    \hfill
    \begin{subfigure}[b]{0.3\textwidth}
        \centering
        \includesvg[width=\textwidth]{downstreamplots/cpt-reloaded-opt-eval__downstream__openbook_qa_len_norm.svg}
        \caption{OpenBookQA \citep{OpenBookQA2018}}
        \label{appendix:downstream:f}
    \end{subfigure}

    \begin{subfigure}[b]{0.3\textwidth}
        \centering
        \includesvg[width=\textwidth]{downstreamplots/cpt-reloaded-opt-eval__downstream__piqa_len_norm.svg}
        \caption{PIQA \citep{Bisk2020}}
        \label{appendix:downstream:g}
    \end{subfigure}
    \hfill
    \begin{subfigure}[b]{0.3\textwidth}
        \centering
        \includesvg[width=\textwidth]{downstreamplots/cpt-reloaded-opt-eval__downstream__rte_len_norm.svg}
        \caption{RTE \citep{dagan2005pascal}}
        \label{appendix:downstream:h}
    \end{subfigure}
    \hfill
    \begin{subfigure}[b]{0.3\textwidth}
        \centering
        \includesvg[width=\textwidth]{downstreamplots/cpt-reloaded-opt-eval__downstream__sciq_acc.svg}
        \caption{SciQ \citep{SciQ}}
        \label{appendix:downstream:i}
    \end{subfigure}

    \begin{subfigure}[b]{0.3\textwidth}
        \centering
        \includesvg[width=\textwidth]{downstreamplots/cpt-reloaded-opt-eval__downstream__sst2_acc.svg}
        \caption{SST-2\citep{socher-etal-2013-recursive}}
        \label{appendix:downstream:j}
    \end{subfigure}
    \hfill
    \begin{subfigure}[b]{0.3\textwidth}
        \centering
        \includesvg[width=\textwidth]{downstreamplots/cpt-reloaded-opt-eval__downstream__winogrande_acc.svg}
        \caption{WinoGrande \citep{ai2:winogrande}}
        \label{appendix:downstream:k}
    \end{subfigure}
    \hfill

\end{figure*}

We report the results of the final downstream evaluation after 10K steps in Table~\ref{tab:final}.
Figure~\ref{appendix:downstream} shows the trajectories of downstream task performance over the course of the training runs.
Notably, full 16-bit training only achieves the best result on the HellaSwag dataset. Full 1.58-bit training achieves the best result on the SciQ dataset.
On all other datasets, one of the continually pre-trained models transitioning into 1.58-bit training at 2K, 4K or 6K steps attains the best downstream evaluation result.
That said, the margin between the downstream results are small and, as expected, the performance of the 16-bit model is on average higher than any specific configuration of the 1.58-bit models. Given the relative volatility of some of the downstream evaluation results and the small differences, it remains unclear to what degree these differences are statistically significant.

\section{Discussion}
Through a comprehensive set of experiments where we have full control over the pre-training data and regimen, we have shown that, counterintuitively, quantization-aware training of 1.58-bit models from scratch is not ideal for obtaining the best possible 1.58-bit models.
In addition, we find that previously proposed strategies of gradually phasing in quantization strength are not needed as long as there are sufficient optimization steps to catch up. 
Furthermore, we make similar observations for the optimizer states: Resetting the optimizer results in a spike in training loss, but the loss can be recovered after relatively few further optimization steps.
Crucially, our experiments on downstream tasks indicate that continual 1.58-bit pre-training is a highly effective training strategy, consistently outperforming full 1.58-bit training, and sometimes even exceeding the performance of full 16-bit training on downstream evaluation tasks.

Our results show that a period of standard training at higher precision (16-bit), and then continuing the pre-training with quantization-aware 1.58-bit training constitutes a preferable training regimen.
We hypothesize that this is because of particular challenges that 1.58-bit quantization-aware training imposes on optimization. Adjusting the weights is more difficult under 1.58-bit training, because the quantized weight may very well remain stable even though the ``shadow weights'' change and their gradient is estimated straight-through. Thus, starting from random parameters, as opposed to a set of parameters obtained through 16-bit training, seems to be more challenging to optimize. Interestingly, the continual pre-training variants with an initial period of standard training have led to lower overall training loss compared to 1.58-bit training from the start, even when being restricted to the same amount of data and number of steps.

This finding has important implications: If one aims for the best possible 1.58-bit model given a fixed amount of data, one should still first train a 16-bit model on a proportion of the data (e.g., between 20 and 40\%) and only then introduce quantization into the training. Keeping the optimizer state is beneficial to avoid loss spikes, but if the optimizer is not available, sufficient further optimization steps enable the model to recover.
Alleviating inference compute demands comes with immediate gains for democratization of AI research~\cite{pmlr-v202-xiao23c} and for advances in scaling test-time compute.

From a different perspective, our results further support the idea of 1.58 continual pre-training being applied to fully pre-trained base models -- opening avenues of research to convert arbitrary 16-bit pre-trained models into corresponding 1.58-bit models. Determining the minimum amount of data required for this conversion is an interesting direction for future work.

We further demonstrate that continually pre-trained 1.58-bit models are competitive on a broad range of downstream tasks, showcasing the capabilities of these inference-efficient models. Notably, the number of parameters was kept fixed between conditions in all our experiments, indicating that those continual 1.58-bit pre-training goes beyond recently proposed scaling laws for quantized networksin general~\cite{kumar2025scaling} and BitNet-style quantization-aware training~\cite{nielsen2024bitnetb158reloadedstateoftheart} in particular. However, an analysis of the scaling behavior of our proposed continual quantization-aware pre-training strategy is left for future work.

\section{Conclusion}

We conducted a systematic comparison between language models trained from scratch with 1.58-bit precision and 16-to-1.58-bit continual pre-training paradigms. Our results show the existence of a data-optimal transition point for switching from 16-bit to 1.58-bit training, providing insights into efficient low-bit training strategies. 
We evaluated the downstream performance of 1.58-bit models, highlighting their viability for real-world applications. These findings contribute to the broader discussion on low-bit training efficiency and its implications for scalable AI model development. Future work may determine the minimal amount of data needed to successfully convert a 16-bit model into a 1.58-bit model through continual pre-training.

\section{Limitations}
The performance on downstream tasks are expected to be further increased when the models undergo supervised instruction fine-tuning and preference optimization before evaluation.  We expect that all models benefit similarly from instruction fine-tuning. However, at this point in time, it cannot be excluded that 16-bit models benefit more from instruction fine-tuning and preference optimizaiton than 1.58-bit models. Future work may specifically investigate the effects of 1.58-bit quantization-aware training during instruction fine-tuning and preference alignment.

\section{Ethical Considerations}
Our work aims at improving the inference compute demands of langauge models. It may contribute to the democratization of AI, alleviating privacy concerns, and to reduce the environmental footprint of LLMs. In particular, our findings suggest that large language models may be converted into lower bit-precision through continual pre-training. We acknowledge that this may come with ethical challenges that govern all research on making powerful models more accessible~\cite{bengio2025international}.
However, this work is purely scientific and does not promote easier access to an actual pool of very powerful models. Using our proposed training strategy would require a similar compute budget as it is needed for standard pre-training.

\bibliography{custom}
\appendix

\end{document}